\documentclass{article} % For LaTeX2e
\usepackage{iclr2019_conference,times}

\usepackage{amsmath,amsfonts,bm}

\def\eqref#1{equation~\ref{#1}}
\def\1{\bm{1}}

\DeclareMathAlphabet{\mathsfit}{\encodingdefault}{\sfdefault}{m}{sl}
\SetMathAlphabet{\mathsfit}{bold}{\encodingdefault}{\sfdefault}{bx}{n}

\DeclareMathOperator*{\argmax}{arg\,max}

\usepackage{algorithm}
\usepackage{algpseudocode}
\usepackage[utf8]{inputenc}
\usepackage[english]{babel}
\usepackage{csquotes}
\usepackage{hyperref}
\usepackage{url}
\usepackage{graphicx}
\usepackage{float}
\usepackage{natbib}

\setcitestyle{numbers}

\title{A Short Survey on \\ Memory based Reinforcement Learning}

\author{Dhruv Ramani  \\
National Institute of Technology, Warangal\\
\texttt{dhruvramani98@gmail.com} 
}

\iclrfinalcopy % Uncomment for camera-ready version, but NOT for submission.
\begin{document}

\maketitle

\begin{abstract}
Reinforcement learning(RL) is a branch of machine learning which is employed to solve various sequential decision making problems without proper supervision. Due to the recent advancement of deep learning, the newly proposed Deep-RL algorithms  have been able to perform extremely well in sophisticated high-dimensional environments. However, even after successes in  many domains, one of the major challenge in these approaches is the high magnitude of interactions with the environment required for efficient decision making. Seeking inspiration from the brain, this problem can be solved by incorporating instance based learning by biasing the decision making on the memories of high rewarding experiences. This paper reviews various recent reinforcement learning methods which incorporate external memory to solve decision making and a survey of them is presented. We provide an overview of the different methods - along with their advantages and disadvantages, applications and the standard experimentation settings used for memory based models. This review hopes to be a helpful resource to provide key insight of the recent advances in the field and provide help in further future development of it.
\end{abstract}

\section{Introduction}
% TODO : Write - Memory is a crucial aspect of an intelligent agent’s ability to plan and reason in partially observable environments. Without memory, agents must act reflexively according only to their immediate per- cepts and cannot execute plans that occur over an extended time interval
% Animals execute goal-directed behaviours despite the limited range and scope of their sensors. To cope, they explore environments and store memories maintaining estimates of important information that is not presently available\cite{clayton1998episodic}.
% Artificial intelligence research is undergoing a renaissance as RL techniques\cite{sutton1998reinforcement}, which address the problem of optimising sequential decisions, have been combined with deep neural networks into artificial agents that can make optimal decisions by processing complex sensory data\cite{mnih2015human}. In tandem, new deep network structures have been developed that encode important prior knowledge for learning problems. One important innovation has been the development of neural networks with external memory systems, allowing computations to be learned that synthesise information from a large number of historical events\cite{weston2014memory,bahdanau2014neural,graves2016hybrid}

Reinforcement Learning (RL) \cite{sutton2018reinforcement} involves an agent to learn to take actions based on it's current situation to maximize a long term reward objective. The actions it takes aren't labeled for training, and the agent has to learn which action to take by trying them all out and settling for the best one. The action it takes has an affect on both the reward it gets immediately and it's long term future. This originates from animal learning in psychology and hence it can mimic human learning ability to select actions that maximize long-term profit in their interactions with the environment. This is the reason why RL has been widely used in robotics and autonomous systems. 

The recent advancement of deep learning has had a significant impact on many areas in machine learning, improving the state-of-the-art in tasks such as object detection \cite{ren2015faster}, speech recognition \cite{graves2013speech}, and language translation \cite{sutskever2014sequence}. The most important property of deep learning is that deep neural networks can automatically find compact low-dimensional representations of high-dimensional data. Deep learning has similarly accelerated progress in RL, with the use of deep neural networks for reinforcement learning tasks, hence defining the field of “deep reinforcement learning”. Deep reinforcement learning involves usage of deep neural networks as a function approximator, thus overcoming the curse of dimensionality. This makes it a promising approach to solving complex real-world problems. 

However, even after the rapid advancements in deep reinforcement learning, the standard architectures are still found to be very sample inefficient and slow. In a setting where the agent learns to play arcade games \cite{mnih2013playing}, deep reinforcement learning systems require millions of interactions with game emulator, amounting to hundreds of hours of game play to achieve human-level performance, which in turn seems pretty inhuman. Addressing the problems in a greater detail, we find that the slow-gradient based update of neural networks requires these algorithms to incur large number of steps to generalize, improve and assimilate the information for policy improvement. For environments with sparse reward signal, modelling the policy with a neural network becomes even more challenging. The low frequency of guiding signals or rewards can be seen as a form of class imbalance where low-reward samples outnumber high-reward samples. Standard reward propagation approaches such as Q-Learning \cite{watkins1992q} cause reward information to be propagated one step at a time through history. However, this flow of information can be fairly efficient if updates happen in reverse order in which the transitions occur. Also, approaches like DQN\cite{mnih2013playing} involve random sampling of experience from the replay memory, to train on uncorrelated mini-batches - requiring the usage of a target network which further slows down this propagation.

% Write more if possible

In order to tackle these shortcomings of the current deep reinforcement learning algorithms, a good idea would be to make decisions based on the experiences which provided high rewards in the past. This involves the agent to be less reflexive according to their immediate perception and make decisions based on the memories it has gathered over an extended time interval. Neither neural network weights, nor activations support storage and retrieval of experiences as the weights change too slowly to store samples as individual experiences. Usage of some variant of recurrent neural networks is common in the partially observable setting \cite{hausknecht2015deep}, however these have trouble learning over long sequences. Also, these approaches do not involve storage of observations as discrete entities, hence the comparison of the new observation with a detailed instance of a rare highly-rewarding past observation becomes very unclear. 

Seeking inspiration from the rapid complementary approach of decision making in the brain \cite{kennerley2011decision}, there have been various recent attempts that try to incorporate external memory modules which enrich the quality of decisions made by the agent (as per the returns accumulated), and make the learning process faster. In the brain, this form of fast learning is supported by the hippocampus and related medial temporal lobe structures \cite{andersen2007hippocampus, thompson1986neurobiology}. Incorporation of this instance based learning strategy serves as a faster rough approximation over a slow generalized decision making system. The recent advancement of memory networks \cite{weston2014memory, sukhbaatar2015end} has attracted a growing amount of interest in the research community to solve the challenging task of designing deep reinforcement learning agents with external memory. In this paper, we provide a review of novel methods which solve this problem. We cover different memory architectures which have been proposed to aid decision making of reinforcement learning agents, different ways in which these proposed architectures are used to solve different sub-problems within reinforcement learning, the environments proposed to test these methods and the applications of these methods. 

% Write more and end properly maybe?

\section{Background}
\label{background}
In this paper we consider a sequential decision making setup, in which an agent interacts with an environment $E$ over discrete time steps. We model the reinforcement learning problem using a Markov Decision Process (MDP), unless otherwise specified. A MDP is defined as a tuple, $\mathcal{M}=(\mathcal{S}, \mathcal{A}, \mathcal{R}, \mathcal{P}, s_0)$, where $\mathcal{S}=\{1,...,s_n\}$ is the state space, $\mathcal{A} = \{1, . . . , a_m\}$ is the action space, and $s_0$ is the initial state distribution. At each time step $t = 1, . . . , \mathcal{T}$ within an episode, the agent observes a state $s_t \in \mathcal{S}$, takes action $a_t \in \mathcal{A}$, receives a reward $r_t \in \mathcal{R}(s_t,a_t)$ and transitions to a new state $s_{t+1} \sim \mathcal{P}(s_t, a_t)$. A reward signal $r \in \mathcal{R}$ is a scalar defining the desirability of an event. A policy $\pi$ is a mapping from a state $s \in \mathcal{S}$ to an action $a \in \mathcal{A}$. 

The agent seeks to maximize the expected discounted return, which is defined as
\begin{equation}
G_t = \sum_{\mathcal{T}=t}^{\infty}\gamma^{\mathcal{T}-t}r_\mathcal{T}
\end{equation}
In this formulation, $\gamma \in [0, 1]$ is a discount factor that trades-off the importance of immediate and future rewards. The state value function $V_\pi (s)$ provides an estimate of the expected amount of the return the agent can accumulate over the future when following a policy $\pi$, starting from any particular state $s$. The action value function $Q_{\pi}(s, a)$ provides the expected return the agent gets when it starts from any particular state $s$, takes an action $a$ and continues to follow the policy $\pi$. 
\begin{gather}
    V_\pi (s) = \mathbb{E}_\pi [G_t | s_t = s] \\
    Q_\pi (s, a) = \mathbb{E}_\pi [G_t | s_t = s, a_t = a]
\end{gather}

$Q^{*}(s,a)$ termed as the optimal action value function is the expected value of taking an action $a$ in state $s$ and then following the optimal policy. In a value based agent, the agent tries to learn an approximation of this and carry out planning and control by acting greedily on it. Q-learning \cite{watkins1992q} is an off-policy control method to find the optimal policy. Q-learning \cite{watkins1992q} uses temporal differences to estimate the value of $Q^{*}(s,a)$. In Q-learning, the agent maintains a table of $Q[\mathcal{S
}, \mathcal{A}]$ and $Q(s, a)$ represents the current estimate of $Q^{*}(s,a)$. This can be learnt by iteratively updating the value using the following :

\begin{equation}
    Q(s, a) \leftarrow Q(s, a) + \alpha [r + \max_{a'} Q(s', a') - Q(s, a)]
\end{equation}

\subsection{Deep Q-Network}
In all the above mentioned equations, the value functions are stored in a tabular form. Because of the memory constraints, lack of generalization and lookup costs, learning to approximate these function is preferred over the tabular setting. Function approximation is a way for generalization when the state and/or action spaces are large or continuous. Function approximation aims to generalize from examples of a function to construct an approximate of the entire function. With the coming of deep learning renaissance, these functions are approximated using neural networks. Deep Q-Network \cite{mnih2013playing} provides the basis of incorporating neural networks to approximate the action value function using Q-learning \cite{watkins1992q} for optimization. In this, the model is parameterized by weights and biases collectively denoted as $\theta$. Q-values are estimated online by querying the output nodes of the network after performing a forward pass  given a state input. Each output unit denotes a separate action. The Q-values are denoted as $Q(s, a |\theta)$ . Instead of updating individual Q-values, updates are now made to the parameters of the network to minimize a differentiable loss function :

\begin{equation}
L(s,a|\theta_i) = \big( r + \gamma \max_{a'} Q(s',a'|\theta_i) - Q(s,a|\theta_i) \big)^2
\end{equation}
\begin{equation}
\theta_{i+1} = \theta_i + \alpha \nabla_\theta L(\theta_i)
\end{equation}

the neural network model naturally generalizes beyond the states and actions it has been trained on. However, because the same network is generating the next state target Q-values that are used in updating its current Q-values, such updates can oscillate or diverge \cite{tsitsiklis1996analysis}. To tackle this problem and to ensure that the neural network doesn't get biased, various techniques are incorporated. Experience Replay is performed in which experiences $e_t = (s_t,a_t,r_t,s_{t+1})$ are recorded in a replay memory $\mathcal{D}$ and then sampled uniformly at training time. This is done to promote generalization. A separate target network $\hat{Q}$ provides update targets to the main network, decoupling the feedback resulting from the network generating its own targets. $\hat{Q}$ is identical to the main network except its parameters $\theta^-$ are updated to match $\theta$ every 10,000 iterations. 

At each training iteration $i$, an experience $e_t = (s_t,a_t,r_t,s_{t+1})$ is sampled uniformly from the replay memory $\mathcal{D}$. The loss of the network is determined as follows:

\begin{equation}
L_i(\theta_i) = \mathbb{E}_{(s_t,a_t,r_t,s_{t+1}) \sim \mathcal{D}} \bigg[ \Big( y_i - Q(s_t,a_t;\theta_i)\Big)^2 \bigg]
\end{equation}

where $y_i = r_t + \gamma \max_{a'}\hat{Q}(s_{t+1},a';\theta^-)$ is the stale update target given by the target network $\hat{Q}$. The actions are taken by acting $\epsilon$-greedily on $Q$. 

The Deep Q-Network has formed the basis of growth for deep reinforcement learning and various modifications have been proposed to make tackle the problems more efficiently. Some of these methods are Double DQN, Prioritized Experience Replay etc.

\subsection{Policy Gradient Algorithms}
The policy $\pi$ is defined as a stochastic or a deterministic function that maps states to the corresponding action which the agent should take . In a policy based agent, the agent tries to learn the policy instead of acting greedily on a value function. Again, in the current times - the policy is represented using a neural network and policy optimization is used to find an optimal mapping. Considering first the vanilla policy gradient algorithm \cite{sutton2000policy}, a stochastic, parameterized policy $\pi_\theta$ is to be optimized. Since we don't know the ground truth labels for the action to take, we go back to maximizing the expected return $J(\pi_\theta) = \mathbb{E}_{\mathcal{T} \sim \pi_\theta} [R(\mathcal{T})]$. The policy is optimized using gradient ascent, to maximize the expected return.

\begin{equation}
\theta_{k+1}=\theta_{k}+\alpha \nabla_{\theta} J\left.\left(\pi_{\theta}\right)\right|_{\theta_{k}}
\end{equation}

where the gradient estimator is of the form 

\begin{equation}
\nabla_{\theta} J(\pi_{\theta}) = \mathbb{E}_{t}\left[\nabla_{\theta} \log \pi_{\theta}\left(a_{t} | s_{t}\right) J(\pi_\theta)\right]
\end{equation}

Majority of the implementations which use automatic differentiation software, optimize the network using an objective whose gradient is equal to the policy gradient estimator, which is obtained by differentiating 

\begin{equation}
L^{P G}(\theta)= \mathbb{E}_{t}\left[\log \pi_{\theta}\left(a_{t} | s_{t}\right) J(\pi_\theta)\right]
\end{equation} 

In order to reduce the variance in the sample estimate for the policy gradient, the return can be replaced by the advantage function $A^{\pi}\left(s_{t}, a_{t}\right)=Q^{\pi}\left(s_{t}, a_{t}\right)-V^{\pi}\left(s_{t}\right)$ which describes how much better or worse it is than other actions on average. 

Among the most popular methods to optimize policies is the Actor-Critic algorithm \cite{konda2000actor}, which learns both a policy and a state-value function, and the value function is used for bootstrapping, i.e., updating a state from subsequent estimates, to reduce variance and accelerate learning. Some of the other commonly used algorithms for policy optimization are TRPO \cite{schulman2015trust}, PPO \cite{schulman2017proximal} and the deterministic \cite{silver2014deterministic} counterparts .

\section{Episodic Memory}
% Write more about episodic memory over here
Episodic control, introduced by  involves learning successful policies based on the episodic memory, which is a key component of human life. In the brain, this form of learning is supported by the hippocampus and related medial temporal lobe structures. Hippocampal learning is thought to be instance-based \cite{marr1991simple, sutherland1989configural}, in contrast to the cortical system which represents generalised statistical summaries of the input distribution \cite{mcclelland1995there}. It is found that humans utilize multiple learning, memory and decision making systems in order to efficiently carry out the task in different situations. For eg. -  when information of the environment is available, the best strategy is model-based planning associated with prefrontal cortex \cite{daw2005uncertainty}. However, when there aren't enough resources to plan, a less intensive decision making system needs to be employed. The common go-to among the RL literature is model-free decision making systems. However, as pointed out earlier - these methods require very high amount of repeated interactions with environment, so there's surely a scope of improvement (when is there not?). This is where episodic control systems come into play and increase the efficiency of model-free control systems by involvement of an external memory system which can represent the hippocampus and inculcate instance based learning in agents. An important observation that supports this approach is that, single experiences with high returns can have prolonged impact on future decision making in humans. For eg. Vasco Da Gamma’s discovery of the sea route to India had an almost immediate and long-lasting effect for the Portuguese. Considering a less extreme eg. - like recalling the plot of a movie as it unfolds, we realize that even in our day to day lives episodic memory plays a crucial role. In reinforcement learning, the experiences of an agent are termed as observations - and in episodic control, the agent leverages the information and advantages of past observations in order to facilitate the decision making progress. For RL agents, \cite{lengyel2008hippocampal}  proposed “episodic control”, which is given as 

\begin{displayquote}
 ... each time the subject experiences a reward that is considered large enough (larger than expected a priori) it stores the specific sequence of State- action pairs leading up to this reward, and tries to follow such a sequence whenever it stumbles upon a State included in it. If multiple successful sequences are available for the same State, the one that yielded maximal reward is followed. 
\end{displayquote}

Majority of the work  which involve episodic memory are based on this. In the next sections, we will go through various models and methods which involve episodic memory for decision making. Majority of the research which involves external memory in RL tasks involves episodic control. Due to this and the biological backdrop to it, this approach for decision making seems promising. 

\section {Memory Modules}
In this section, we go through various papers which have proposed a new memory module/method which can be used for episodic control. All of the methods presented below propose a novel method to incorporate memory into the reinforcement learning setting. These modules can be used to solve various different kind of problems in reinforcement learning.

\subsection{Model Free Episodic Control}
This \cite{blundell2016model} can be considered one of the first few works which involved external memory in the reinforcement learning scene. It considers a deterministic environment, given the near deterministic situations in the real world and the specialised learning mechanisms in the brain which exploit this structure. The episodic model which represents the hippocampus for instance based learning here is represented by a non-parametric growing table which is indexed by state-actions pair. It is represented by $Q^{EC}(s, a)$ - and is used to rapidly record and replay the sequence of actions that so far yielded the highest return from a given start State. Size of the table is limited by removing the least recently updated entry once the maximum size of the table is reached.

At the end of each episode, the table is updated in the following way :

\begin{eqnarray}
    \label{eq:qec}
    Q^{EC}(s_t, a_t) &\leftarrow
    \begin{cases}
        R_t &\text{if $(s_t, a_t) \not\in Q^{EC}$,} \\
        \max \left\{Q^{EC}(s_t, a_t), R_t\right \} &\text{otherwise}
    \end{cases}
\end{eqnarray}
Thus, the values stored in $Q^{EC}(s, a)$  do not correspond to estimates of the expected return, rather they are estimates of the highest potential return for a given State and action.

The value is estimated in the following way :

\begin{eqnarray}
    \label{eq:qecknn}
    \widehat{Q^{EC}}(s,a) = 
    \begin{cases}
        \frac{1}{k} \sum_{i=1}^k Q^{EC}(s^{(i)}, a) & \text{if $(s,a) \not\in Q^{EC}$,} \\
        Q^{EC}(s,a) & \text{otherwise}
    \end{cases}
\end{eqnarray}

For the states which have never been visited, $Q^{EC}(s, a)$ is approximated by taking average of the K-nearest states.

The main algorithm is given as follows : 
\begin{algorithm}[h]
    \caption{Model-Free Episodic Control.
		\label{alg:mfeccap}}
	\begin{algorithmic}
		\For{each episode}
            \For{$t = 1, 2, 3, \dots, T$}
                \State Receive observation $o_t$ from environment.
                \State Let $s_t = \phi(o_t)$.
                \State Estimate return for each action $a$ via (\eqref{eq:qecknn})
                \State Let $a_t = \arg\max_a \widehat{Q^{EC}}(s_t, a)$
                \State Take action $a_t$, receive reward $r_{t+1}$
		    \EndFor
		    \For{$t = T,T-1,\ldots,1$}
                \State Update $Q^{EC}(s_t, a_t)$ using $R_t$ according to (\eqref{eq:qec}).
		    \EndFor
		\EndFor
	\end{algorithmic}
\end{algorithm}

Here, $\phi$ is a feature mapping which maps the observation $o_t$ to the state $s_t$. In this paper, $\phi$ is represented as a projection to smaller-dimensional space ie. $\phi:x \to \mathcal{A}x$, where $\mathcal{A} \in \mathbb{R}^{F \times D}$ and $F \ll D$ where $D$ is the dimensionality of the observation and $\mathcal{A}$ is a random matrix drawn from a standard Gaussian. $\phi$ can also be represented as a latent-variable probabilistic models such as a variational autoencoder \cite{kingma2013auto}.

This approach had been the Arcade Learning Environment (Atari) \cite{mnih2013playing}, and a first-person 3-dimensional environment, and it was found that it provided faster results as compared to standard function approximators.

\subsection{Neural Episodic Control}
\begin{figure}[h]
    \centering
    \includegraphics[width=15cm]{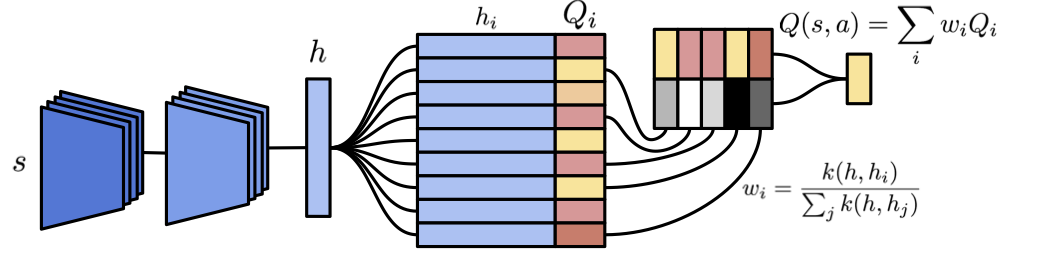}
    \caption{\textbf{Neural Episodic Control} - Architecture of episodic memory module for a single action $a$. Pixels representing the current state enter through a convolutional neural network on the bottom left and
    an estimate of $Q(s,a)$ exits top right. Gradients flow through the entire architecture.}
    \label{fig:arch-dnd}
\end{figure}

This \cite{pritzel2017neural} was the first end-to-end architecture which involved using memory in RL and was entirely differentiable. The agent in this method consists of three components - a DQN \cite{mnih2013playing} inspired convolutional neural network that processes pixel images $s$ and brings it down to an embedding space $h$. This is then used to index a set of memory modules (one per action). These readouts from the action memories are converted to $Q(s, a)$ using a final network. Figure~\ref{fig:arch-dnd} shows the architecture during a single pass.

The memory architecture in this paper is defined as a Differential Neural Dictionary (DND) - in which each action $a\in\mathcal{A}$ has a simple memory module $M_a = (K_a, V_a)$, where $K_a$ and $V_a$ are dynamically sized arrays of vectors, each containing the same number of vectors. The keys here correspond to the embedding $h$ which represent the state $s$ and the value correspond to the respective $Q(s, a)$ values. For each key $h_i$, the lookup from the dictionary is performed as a weighted summation of the P-nearest corresponding values. 

\begin{align}
\label{eq:outweight}
o &= \sum_i^p  w_i v_i,
\end{align}
where $v_i$ is the $i$th element of the array $V_a$ and
\begin{equation}
    w_i = k(h, h_i)/\sum_j k(h, h_j),
    \label{eq:weighted-sum}    
\end{equation}
Here, $k$ is represented as a Gaussian or an inverse kernel. In the experiments performed, $k$ is taken as :
\begin{equation}
    k(h, h_i) = \frac{1}{\|h-h_i\|_2^2 + \delta}.
\end{equation}

The writes to the DND are append only, and if the key already exits it's updated. The values are updated by performing standard Q-learning \cite{watkins1992q} updates with $N$-step Q-learning \cite{peng1994incremental}. 
\begin{align}
    Q^{(N)}(s_t,a) &= \sum_{j=0}^{N-1} \gamma^j r_{t+j} + \gamma^N \max_{a'} Q(s_{t+N}, a') \\
    \label{eq:nstepq}
Q_i &\leftarrow Q_i + \alpha (Q^{(N)}(s,a) - Q_i) 
%\label{eq:fast-updates}
\end{align}
where $\alpha$ is the learning rate of the $Q$ update. 

% TODO - Fix this

The algorithm for NEC is given at \ref{alg:neccap}. 

\begin{algorithm}[h]
    \caption{Neural Episodic Control
        \label{alg:neccap}}
    \begin{algorithmic}
        \State $\mathcal{D}$: replay memory.
        \State $M_a$: a DND for each action $a$.
        \State $N$: horizon for $N$-step $Q$ estimate.
        \For{each episode}
            \For{$t = 1, 2, \dots, T$}
                \State Receive observation $s_t$ from environment with embedding $h$.
                \State Estimate $Q(s_t, a)$ for each action $a$ via \eqref{eq:outweight} from $M_a$
                \State $a_t \leftarrow$ $\epsilon$-greedy policy based on $Q(s_t,a)$
                \State Take action $a_t$, receive reward $r_{t+1}$
                \State Append $(h,Q^{(N)}(s_t,a_t))$ to $M_{a_t}$.
                \State Append $(s_t, a_t, Q^{(N)}(s_t,a_t))$ to $\mathcal{D}$.
                \State Train on a random minibatch from $\mathcal{D}$.
            \EndFor
        \EndFor
    \end{algorithmic}
    \label{alg:nec}
\end{algorithm}

\subsection{MASKED EXPERIENCE MEMORY}
This architecture proposed in \cite{loynd2018now} provides a new memory module $M$ which uses mask vectors in the read operation which provides distributed weightage to the past memories based on the current observations for control. The write operation appends the last observation into the fixed size memory store, while the oldest memory is dropped from the store. The read operation compares the current observation with all the previously written observations in the memory store and returns a vector calculated as a weighted sum of all memories. 

\begin{equation}
    R = \sum_{i=1}^N b_iM_i
\end{equation}
$N$ is the number of $D$ dimension vectors in the memory module and $R$ is called the read vector. $R$ is defined as the weighted ($b$) summation over memories similarity ($Q_i$) to the current read vector.
\begin{equation}
    b_i = \frac{\exp(Q_i)}{\sum_{j=1}^N \exp(Q_j)}
\end{equation}

 $Q_i$ is defined as the masked summation of the euclidean distance between the current observation/read key $s$ and the memory elements : 
\begin{equation}
    Q_i = \exp(z)\sum_{d=1}^D a_d (s - M_{id})^2
\end{equation}
\begin{equation}
    a_d = \frac{\exp(w_d)}{\sum_{k=1}^D \exp(w_k)}
\end{equation}

 the mask weight vector $w$ and attention distribution sharpness parameter $z$ are trained by gradient descent. The architecture used is a LSTM \cite{hochreiter1997long} based actor-critic \cite{konda2000actor} networks. Before every episode, the memory is cleared. At every time-step, the current observation, the one-hot representation of the last action taken, the reward received and the memory readout (from the current observation) is concatenated and passed as an input to both, the actor-LSTM and the critic-LSTM. The standard policy-gradient training procedure is followed. The key contribution of this architecture is the usage of attention \cite{vaswani2017attention} mechanism which causes the agent to learn which memory segments to focus on. This paper also proposed novel tasks to test memory-based frameworks on (which are discussed later).

\subsection{Integrating Episodic Memory with Reservoir Sampling}
% NOTE : EDIT!

This \cite{j.2018integrating} method introduces an end-to-end trainable episodic memory module. Instead of assigning credit to the recorded state by explicitly backpropogating through time, the set of states from the memory are drawn from a distribution over all n-subsets of visited states which are parameterized by weights. To draw from such a distribution without maintaining all visited states in memory, a reservoir sampling technique is used. 

The model is based on advantage actor critic \cite{konda2000actor} architecture, consisting of separate value and policy networks. In addition to that, there's an external memory module $\mathbb{M}$ consisting of $n$ past visited states $(S_{t_0}, ..., S_{t_{n-1}})$ with associated important weights $(w_{t_0}, ..., w_{t_{n-1}}$. Other than this, other trainable networks include a query network ($q$), write network ($w$). The state $S_t$ is given separately to the query, write, value and networks at each time step.  The query network outputs a vector of size equal to the input state size which is used to choose a past state from the memory, which is taken as an input by the policy. The write network assigns a weight to each new state determining how likely it is to stay in memory. The policy network assigns probabilities to each action conditioned on current state and recalled state. The value network estimates expected return (value) from the current state.

\begin{equation}
    \label{eq:15}
    m_t = Q(S_{t_i} | \mathcal{M}_t) = \frac{\exp(\frac{(q(S_t) | S_{t_i})}{\mathcal{T}})}{\sum_{j=0}^{n-1} \exp(\frac{(q(S_t) | S_{t_j})}{\mathcal{T}})}
\end{equation}

The model is trained using stochastic gradient descent, using standard RL loss functions for actor-critic \cite{konda2000actor} method. The query network is trained on the loss $-\delta_t \log(Q(m_t | S_t))$, by freezing the other networks. With the write network, the weights $w(S_t)$ learned by the network are used in a reservoir sampling algorithm such that the probability of a particular state $S_t$ being in the memory at given future time is proportional to associated weights $w(S_t)$ and to obtain estimates of the gradient of the return with respect to the weight. 

For sampling from the memory, a distribution with respect to the weights is learnt. The expected return is calculated based on this distribution and the policy and other networks are trained using policy gradient algorithm. However, sampling the samples from this distributions involve usage of a reservoir sampling \cite{vitter1985random} algorithm which allows a steady distribution even while the contents of the memory change. The sampling algorithm is too big to covered in this review, and can be read in the paper.

\subsection{Neural Map}
The proposed Neural Map \cite{parisotto2017neural} memory module was specifically designed for episodic decision making in 3D environments, under a partial observable setting. This method provides a lot of weightage on the area where the agent is currently located for the storage of memory and decision making. The write operator for the memory is selectively limited to affect the part of the neural map which represents agent's current position. If the position of the agent is given by $(x, y)$ with $x \in\mathbb{R}$ and $y \in\mathbb{R}$ and the neural map $M$ is a $C\times H\times W$ feature block, where $C$ is the feature dimension, $H$ is the vertical extent of the map and $W$ is the horizontal extent. There exists a coordinate normalization function $\psi(x,y)$ which maps every unique $(x,y)$ to $(x',y')$, where $x'\in \{0,\hdots,W\}$ and $y'\in \{0,\hdots,H\}$. 

\begin{figure}[h]
    \centering
    \includegraphics[width=15cm]{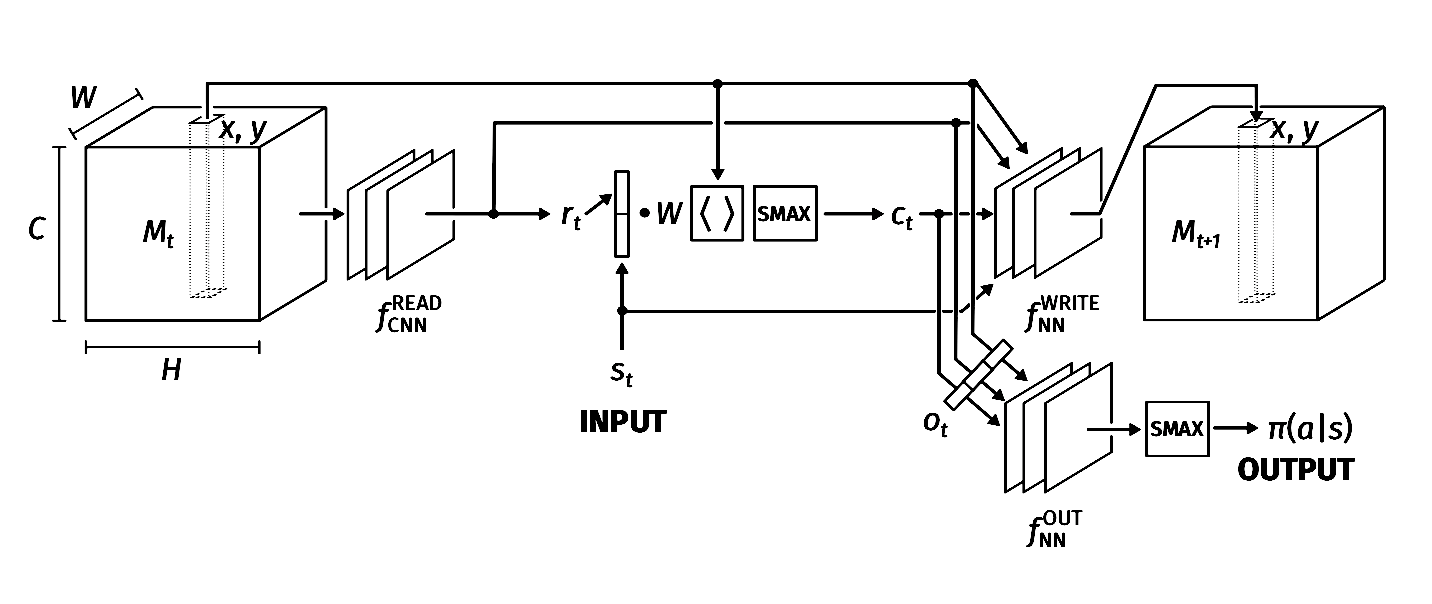}
    \caption{\cite{uwaterlooslides} \textbf{Neural Map} - The method involves operation on the external memory $M_t$ based on the current position of the agent within the environment.}
    \label{fig:arch-nm}
\end{figure}

Let $s_t$ be the current state embedding, $M_t$ be the current neural map, and $(x_t,y_t)$ be the current position of the agent within the neural map. 
The Neural Map is defined by the following set of equations:
\begin{align}
  r_t                 &= read(M_t) \\
  c_t                 &= context(M_t, s_t, r_t) \\
  w_{t+1}^{(x_t,y_t)} &= write(s_t, r_t, c_t, M_t^{(x_t,y_t)}) \\
  M_{t+1}             &= update(M_t, w_{t+1}^{(x_t,y_t)}) \\
  o_t                 &= [r_t, c_t, w_{t+1}^{(x_t,y_t)}] \\
  \pi_t(a|s)          &= \text{Softmax}(f(o_t)),
\end{align}
where 
$w_{t}^{(x_t,y_t)}$ represents the feature at position $(x_t,y_t)$ at time $t$,
$[x_1,\hdots,x_k]$ represents a concatenation operation, and 
$o_{t}$ is the output of the neural map at time $t$ which is then processed by another deep network $f$ to get the policy outputs $\pi_t(a|s)$. The global $read$ operation produces a $C$-dimensional feature vector $r_t$ by passing the neural map $M_t$ through a deep convolutional network, hence summarizing the current instance of the memory. The $context$ read operation is used to check for certain features in the map and is given as : 
\begin{align}
  \label{eq:context1}
  q_t              &= W [s_t, r_t] \\
  \label{eq:context2}
  a_t^{(x,y)}      &= q_t \cdot M_t^{(x,y)} \\
  \label{eq:contextprob}
  \alpha_t^{(x,y)} &= \frac{e^{a_t^{(x,y)}}}{\sum_{(w,z)} e^{a_t^{(w,z)}}} \\
  \label{eq:context3}
  c_t              &= \sum_{(x,y)} \alpha_t^{(x,y)} M_t^{(x,y)}, 
\end{align}

Where $s_t$ is the current state embedding, $r_t$ is the current global read vector and they first produce a query vector $q_t$. The inner product of the query vector and each feature $M_t^{(x,y)}$ in the neural map is then taken to get scores $a_t^{(x,y)}$ at all positions $(x,y)$. Soft attention \cite{vaswani2017attention} is appplied over these scores to get the context vector $c_t$. Based on the representations calculated above and the current coordinates of the agent, the write operation is performed using a deep neural network $f$ which gives a new C-dimensional write candidate vector at the current position $(x_t,y_t)$  
\begin{align}
  w_{t+1}^{(x_t,y_t)} &= f([s_t, r_t, c_t, M_t^{(x_t,y_t)}])
\end{align}
This write vector is used to update the memory in the following way : 
\begin{align}
  M_{t+1}^{(a,b)} = \left\{ \begin{array}{lr}
    w_{t+1}^{(x_t,y_t)}, & \text{for } (a,b)=(x_t,y_t) \\
    M_t^{(a,b)},         & \text{for } (a,b)\neq(x_t,y_t)
  \end{array}\right.
\end{align}

For experimentation, the agent had been trained on a 3D Maze environment, where the only observation given was the current forward view of the agent. The memory represents a 2D map of the whole maze. The visualized activations for the mapping provided key insights about the rewarding trajectories which the agent should take.

\subsection{Memory, RL, and Inference Network (MERLIN)}
This \cite{wayne2018unsupervised} seminal paper combined external memory systems, reinforcement learning and variational inference \cite{kingma2013auto} over states into a unified system based on concepts from psychology and neuroscience - predictive sensory coding, the hippocampal representation theory of Gluck and Myers\cite{gluck1993hippocampal}, and the temporal context model and successor representation. Information from various sensory input modalities (image $I_t$, egocentric velocity $v_t$, previous reward $r_{t-1}$ and action $a_{t-1}$, and a text instruction $T_t$) are taken as observation $o_t$ and are encoded to $e_t = \text{enc}(o_t)$. All of these encoders were taken as ResNet \cite{he2016deep} modules. 

\begin{figure}[h!]
    \centering
    \includegraphics[width=15cm]{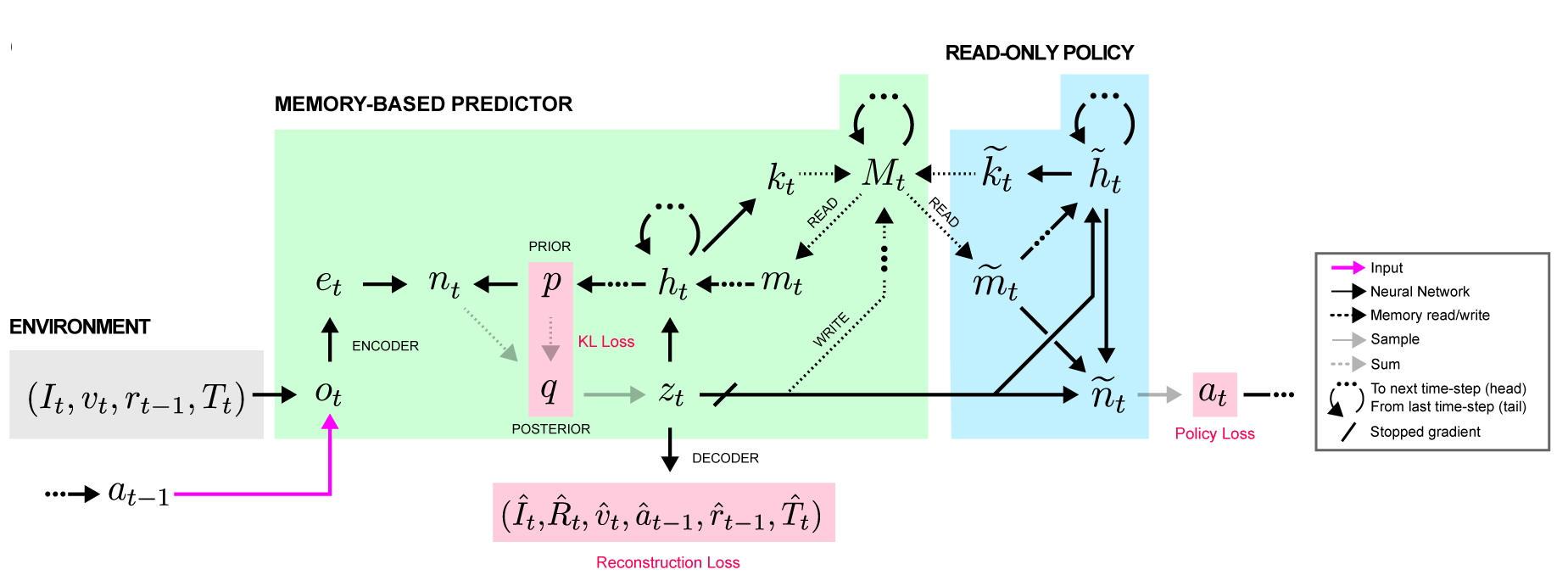}
    \caption{\textbf{MERLIN} - The architecture consists of two sub-networks, one for policy prediction and other for observation inference. Both use a common external memory $M_t$ and are modelled by a recurrent neural network.} 
    \label{fig:arch-merlin}
\end{figure}

Based on variational inference, which treats inference as an optimization problem - the model considers a \emph{prior} distribution which predicts the next state variable conditioned on a history maintained in memory of the previous state variables and actions: $p(z_{t} | z_1, a_1, \dots, z_{t-1}, a_{t-1})$. The \emph{posterior} distribution corrects this prior based on the new observations $o_t$ to form a better estimate of the state variable: $q(z_t | z_1, a_1, \dots, z_{t-1}, a_{t-1}, o_t)$. The mean and log standard deviation of the prior distribution $p$ are concatenated with the embedding and passed through a network to form an intermediate variable $n_t$, which is added to the prior to make a Gaussian posterior distribution $q$, from which the state variable $z_t$ is sampled. This is inserted into row $t$ of the memory matrix $M_t$ and passed to the recurrent network $h_t$. The memory is represented using a Differentiable Neural Computer (DNC) \cite{graves2016hybrid} and the recurrent network is represented by a deep LSTM \cite{hochreiter1997long}. The recurrent network has several read heads each with a key $k_t$, which is used to find matching items $m_t$ in memory. The state variable is passed as input to the read-only policy and is passed through decoders that produce reconstructed input data ($\hat{I_t}$, $\hat{v_t}$, $\hat{r_{t-1}}$ and $\hat{a_{t-1}}$) and the return prediction $\hat{R}_t$. These decoders were taken as dual of the respective encoders with transposed convolutions wherever required. The MBP is to be optimized to produce predictions that are consistent with the probabilities of observed sensory sequences from the environment: $\Pr(o_1,o_2,\dots)$. This objective can be intractable, hence based on the standard variational inference procedure, the MBP is trained instead to optimise the variational lower bound (VLB) loss, which acts as a tractable surrogate.
\begin{align}
\log p(o_{0:t}, R_{0:t}) & \geq \sum_{\tau=0}^t \mathbb{E}_{q(z_{0:\tau-1} | o_{0:\tau-1})} \bigg[\mathbb{E}_{q(z_{\tau} | z_{0:\tau-1}, o_{0:\tau})} \log p(o_\tau, R_\tau | z_\tau) \nonumber \\ & \hspace{1cm}- \text{D}_\text{KL}[q(z_\tau | z_{0:\tau-1}, o_{0:\tau}) || p(z_\tau | z_{0:\tau-1}, a_{0:\tau-1})] \bigg]. 
\label{eq:return_vlb}
\end{align}

This loss consists of a reconstruction loss and a KL divergence between $p$ and $q$. To implement the reconstruction term, several decoder networks take $z_t$ as input, and each one transforms back into the space of a sensory modality. The difference between the decoder outputs and the ground truth data is the loss term. The KL divergence between the prior and posterior probability distributions ensures that the predictive prior is consistent with the posterior produced after observing new sensory inputs.

The policy, which has read-only access to the memory, is the only part of the system that is trained conventionally according to standard policy gradient algorithms. To emphasise the independence of the policy from the MBP, the gradients are blocked from the policy loss into the MBP.

MERLIN also excelled at solving one-shot navigation problems from raw sensory input in randomly generated, partially observed 3D environments. It had also effectively learned the ability to locate a goal in a novel environment map and quickly return to it. Even though, it wasn't explicitly programmed, MERLIN showed evidence of hierarchical goal-directed behaviour, which was detected from the MBP's read operations. 

\subsection{Memory Augmented Control Network}

This model \cite{khan2017memory} was specifically created to target partially observable environment with sparse rewards, both being a common and tricky problems in reinforcement learning because of lack of immediate feedback leading to hard to model navigation policies. Similar to the options framework \cite{stolle2002learning} in hierarchical reinforcement learning, this architecture breaks the planning problem into two levels. At a lower level, a planning module computes optimal policies using a feature rich representation of the locally observed environment. The higher level policy is used to augment the neural memory to produce an optimal policy for the global environment. In the set of experiments, the agent operates in an unknown environment and must remain safe by avoiding collisions. Let $m \in \{-1,0\}^n$ be a \emph{hidden} labeling of the states into free $(0)$ and occupied $(-1)$. The agent has access to a sensor that reveals the labeling of nearby states through an observations $z_t = H(s_t)m \in \{-1,0\}^{n}$, where $H(s) \in \mathbb{R}^{n \times n}$ captures the local field of view of the agent at state $s$. The agent's task is to reach a goal region $\mathcal{S}^{\text{goal}} \subset \mathcal{S}$, which is assumed obstacle-free, i.e., $m[s] = 0$ for all $s \in \mathcal{S}^{\text{goal}}$. The information available to the agent at time $t$ to compute its action $a_t$ is $h_t := (s_{0:t}, z_{0:t}, a_{0:t-1}, \mathcal{S}^{\text{goal}}) \in \mathcal{H}$, where $\mathcal{H}$ is the set of possible sequences of observations, states, and actions. A policy $\mu : \mathcal{S} \rightarrow \mathcal{A}$ is to be learnt such that the agent is able to reach the goal state without any obstacles in between. The partial observability requires consideration of \emph{memory} in order to learn $\mu$ successfully.

\begin{figure}[h]
  \centering
  \includegraphics[width = 15cm]{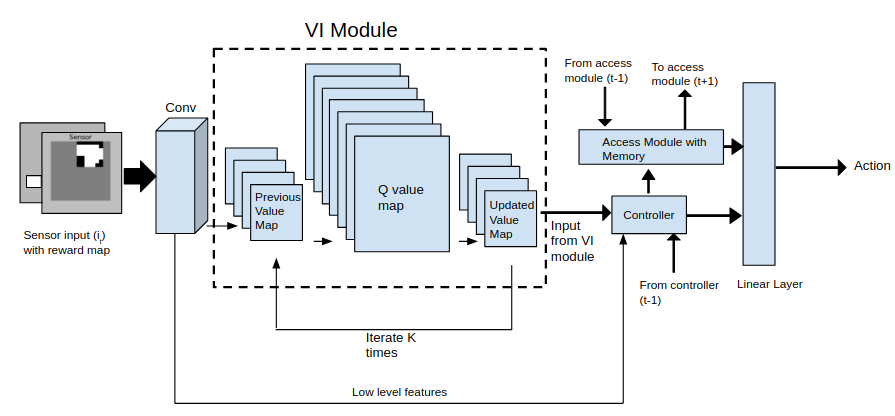}
  \caption{\textbf{MACN} - The architecture uses convolutional layers to extract features from the environment. The value maps are generated with these features. The controller network uses the value maps and low level features to emit read and write heads in addition to doing its own planning computation. \label{fig:MACN}}
\end{figure}

A partially observable markov decision process based on the history space $\mathcal{H}$ is defined by $\mathcal{M}(\mathcal{H},\mathcal{A},\mathcal{T},r,\gamma)$, where $\gamma \in (0,1]$ is a discount factor, $\mathcal{T}: \mathcal{H} \times \mathcal{A} \rightarrow \mathcal{H}$ is a deterministic transition function, and $r: \mathcal{H} \rightarrow \mathbb{R}$ is the reward function, defined as follows:
\begin{align}
  \mathcal{T}(h_t,a_t) &= (h_t,s_{t+1} = f(s_t,a_t), z_{t+1} = H(s_{t+1})m, a_t)\\
  r(h_t,a_t) &= z_t[s_t]
\end{align}
At any instance, the agent observes a small part of the environment (local observed space). This approach computes optimal policies for these locally observed spaces and then uses these to compute a policy optimal in the global space. The read and write operators on the memory $M_t$ are defined as 
\begin{align}
    {re}_{t}^i &= M_{t}^{\top}w_{t}^{read,i} \\
     M_{t} &=  M_{t-1} (1 - w_{t}^W e_{t}^\top) + w_{t}^Wv_{t}^\top\enspace
\end{align}
where $w_{t}^{read,i}$ are the read weights, $w_{t}^W$ are write weights, $e_{t}$ is an erase vector and $v_{t}$ is a write vector. The write vector and the erase vector are emitted by the controller. At a lower level, planning is done in a local space given by $z'$ within the boundaries of our locally observed environment space. This setting can be formulated as a fully observable markov decision process given by $\mathcal{M}_t(\mathcal{S},\mathcal{A},f,r,\gamma)$ and planning in this is done by calculating an optimal policy for this local space given by $\pi_{l}^*$.  Let $\Pi=[\pi_{l}^{1},\pi_{l}^{2},\pi_{l}^{3},\pi_{l}^{4},\dots,\pi_{l}^{n}]$ be the list of optimal policies calculated from such consecutive observation spaces [$z_0,z_1,\dots z_T$]. These two are mapped by training a convolutional neural network using standard policy gradient approaches. 

Hence, in this model a value iteration network \cite{tamar2016value} is used to learn the value maps of the observation space $z$. These value maps are used as keys for the differential memory, and are found to perform better planning than just standard CNN embeddings. These local value maps (used to calculate local policies) are concatenated with a low level feature representation of the environment and sent to a controller network. The controller network interfaces with the memory through an access module (another network layer) and emits read heads, write heads and access heads. In addition, the controller network also performs its own computation for planning. The output from the controller network and the access module are concatenated and sent through a linear layer to produce an action. This entire architecture is then trained end to end.

\section{Usage of Episodic Memory}
The memory architectures presented above have been used to carry out various different tasks in reinforcement learning. Majority of the literature surveyed involves usage of external memory systems to make the existing deep-learning based algorithms more efficient, in terms of the number of interactions, enhancing reward propagation and having strong priors from the past for decision making. In this section, we go through various algorithms which don't propose a new memory module - however, are heavily depend upon the usage of episodic memory to accomplish the respective task.

\subsection{Episodic Backward Update}
This method \cite{lee2018sample} involves episodic memory to train a Deep Q-Network \cite{mnih2013playing} with backward updates through time. It is based on the observation that whenever we observe an event, we scan through our memory in a backward manner and recognize the relationships between the current observation and the past experiences \cite{lengyel2008hippocampal}. The simple backward update of the Q value function in the tabular setting - which first generates the entire episode and performs backward update as defined in algorithm \ref{alg:simpleback} is very unstable if applied to deep reinforcement learning settings.  

 \begin{algorithm}[h]
   \caption{Simple Episodic Backward Update (single episode, tabular)}
   \label{alg:simpleback}
    \begin{algorithmic}
      \State Initialize the Q- table $Q \in \mathbb{R}^{\mathcal{S} \times \mathcal{A}}$ with zero matrix. \newline $Q(s,a) = 0$ for all state action pairs $(s,a) \in \mathcal{S} \times \mathcal{A}$.
      \State Experience an episode $ E = \{(s_1,a_1,r_1,s_2), \ldots, (s_{T},a_{T},r_{T},s_{T+1})\}$
      \For {$t = T$ to 1}
          \State $Q(s_t,a_t) \leftarrow r_t + \gamma \max_{a^\prime} Q(s_{t+1},a')$

      \EndFor

\end{algorithmic}
\end{algorithm}

Hence, this algorithm is modified (\ref{alg:back2}) to perform backward updates in the deep-learning setting by using episodic memory. All the transitions within the samples epidsode $E = \{\bm{S,A,R,S^\prime}\}$ are used, where $E$ is denoted as a set of four length-$T$ vectors: $ \bm{S}=\{S_1, S_2, \ldots S_T\}$; $\bm{A}=\{A_1,A_2,\ldots A_T\}$; $\bm{R}=\{R_1,R_2,\ldots R_T\}$ and $\bm{S^\prime}=\{S_2, S_3,\ldots S_{T+1}\}$. Episodic memory based module is used as temporary target - $\tilde{Q}$ and it is initialized to store all the target Q-values of $\bm{S^\prime}$ for all valid actions. $\tilde{Q}$ is an $|\mathcal{A}|\times T$ matrix which stores the target Q-values of all states $\bm{S^\prime}$ for all valid actions. Therefore, the $j$-th column of $\tilde{Q}$ is a column vector that contains $\hat{Q}\left(S_{j+1},a;\bm{\theta^-}\right)$ for all valid actions $a$, where $\hat{Q}$ is the target Q-function parameterized by $\theta^-$. 

\begin{algorithm}[h]
   \caption{Episodic Backward Update}
   \label{alg:back2}
    \begin{algorithmic}
      \State Initialize replay memory $D$ to capacity $N$
      \State Initialize on-line action-value function $Q$ with random weights $\bm{\theta}$
      \State Initialize target action-value function $\hat{Q}$ with random weights $\bm{\theta^-}$
      \For {episode = 1 to $M$ }
        \For {$t = 1$ to \mbox{Terminal} } 

          \State With probability $\epsilon$ select a random action $a_t$
          \State Otherwise select $a_t = \argmax_a Q\left(s_t,a;\bm{\theta}\right)$     
          \State Execute action $a_t$, observe reward $r_t$ and next state $s_{t+1}$
          \State Store transition $(s_t,a_t,r_t,s_{t+1})$ in $D$

          \State Sample a random episode $E = \{\bm{S,A,R,S^\prime}\}$ from $D$, set $T=\operatorname{length}(E)$

          \State Generate temporary target Q table, $\tilde{Q} = \hat{Q}\left(\bm{S^\prime},\bm{\cdot};\bm{\theta^-}\right)$
          \State Initialize target vector $\bm{y} =  \operatorname{zeros}(T)$

          \State $\bm{y}_T \leftarrow \bm{R}_T$
          \For {$k = T-1$ to 1}
            \State $\tilde{Q}\left[\bm{A}_{k+1}, k\right] \leftarrow \beta \bm{y}_{k+1} + (1-\beta) \tilde{Q}\left[\bm{A}_{k+1}, k\right]$
            \State $\bm{y}_k \leftarrow  \bm{R}_k + \gamma\max_{a\in\mathcal{A}}  \tilde{Q}\left[a, k\right] $
          \EndFor

          \State Perform a gradient descent step on $\left(\bm{y}-Q\left(\bm{S},\bm{A};\bm{\theta}\right)\right)^2$ with respect to $\bm{\theta}$
          \State Every C steps reset $\hat{Q} = Q$

        \EndFor
      \EndFor
    \end{algorithmic}
\end{algorithm}

The target vector $\bm{y}$ is used to train the network by minimizing the loss between each $Q\left(S_j,A_j;\theta\right)$ and $\bm{y}_j$ for all $j$ from 1 to $T$. Adopting the backward update idea, one element $\tilde{Q}\left[\bm{A}_{k+1},k\right]$ in the $k$-th column of the $\tilde{Q}$ is replaced using the next transition's target $y_{k+1}$. Then $y_k$ is estimated as the maximum value of the newly modified $k$-th column of $\tilde{Q}$. This procedure is repeated in a recursive manner and the backward update is finally applied to Deep Q-networks \cite{mnih2013playing}. This algorithm has been tested 2D maze environment and the Arcade Learning Environment \cite{mnih2013playing} and provides a novel way to perform backward updates on deep architectures using episodic memory. 

\subsection{Episodic Memory Deep Q-Networks}

This method \cite{lin2018episodic} tries to improve the efficiency of DQN \cite{mnih2013playing} by incorporating episodic memory - mimicking the competitive and cooperative relationship between Striattum and Hippocampus in the brain. This approach combines the generalization strength of DQN \cite{mnih2013playing} and the fast converging property of episodic memory, by distilling the information in the memory to the parametric model. The DQN function is paramterized by $\theta$ and is represented by $Q_\theta$, while the episodic memory targets are represented by $H$, given by :

\begin{equation} 
    H(s_t, a_t) = \max_i R_i(s_t, a_t), i \in (1, 2, \cdots, E)
\end{equation}

where $E$ represents the number of episodes that the agent has experienced, and $R_i(s, a)$ represents future return when taking action $a$ under state $s$ in i-th episode. $H$ is a growing table indexed by state-action pairs $(s, a)$ and is implemented in a way similar to \cite{}. % Model free episodic memory
The loss function given below is minimized to train $Q_\theta$ : 

\begin{equation} 
    L = \alpha(r_t + \gamma \max_{a'} Q_\theta(s_{t+1}, a') - Q_\theta)^{2} + \beta (H - Q_\theta)^2
\end{equation}

Though straight-forward, this paper has reported various advantages over vanilla-DQN. Because the memory stores optimal rewards, the reward propagation through the network is faster, compensating the disadvantage of slow-learning resulted by single step reward update. Introducing the memory module also makes this DQN highly sample efficient. This architecture had been tested on the Atari suite \cite{mnih2013playing} and had significantly outperformed the original model.

\subsection{Episodic Curiosity through Reachability}
This paper \cite{savinov2018episodic} specifically address the environments with sparse rewards. This is unsurprisingly common, most of the environments provide no or negative rewards for non-final states, and a positive reward for the final state. Due to very infrequent supervision signal, a common trend among researchers is to introduce rewards which are internal to the agent and is thus called Intrinsic Motivation \cite{chentanez2005intrinsically} or Curiosity Driven Learning \cite{pathak2017curiosity}. 

This paper provides an internal reward when it reaches specific states which is non-final, and is considered as novel as per the method. The states which require effort to reach (based on the number of environment steps taken to reach it are considered as novel. To estimate this, a neural network $C(E(o_i), E(o_j))$ is trained to predict the number of steps that separate two observations. This is binarized, hence the network predicts a value close to $0$ if the number of steps that separate them is less than $k$, which is a hyper-parameter. $E(o)$ is also a neural network, which brings the observation $o$ to a lower embedding. Among the two observations $(o_i, o_j)$ - one is the current observation and the second is a roll-out from an episodic memory bank which stores the past observations. If the predicted number of steps between these two observations is greater than a certain threshold, the agent rewards itself with a bonus, and adds this observation to the episodic memory. The episodic memory $M$ of size $K$ stores the embeddings of the past observations. At every time step, the current observation o goes through the embedding network producing the embedding vector $e = E(o)$. This embedding vector is compared with the stored embeddings in the memory buffer $M = e_1 , \hdots, e_{|M|}$ via the comparator network $C$ where $|M|$ is the current number of elements in memory. This comparator network fills the reachability buffer with values
\begin{equation}
c_i = C(e_i, e), i = 1, \hdots |M|
\end{equation}
Then the similarity score between the memory buffer and the current embedding is computed as 
\begin{equation}
C(M,e) = F(c_1, ..., c_{|M|})\in [0,1]
\end{equation}
The internal reward, called the curiosity bonus is calculated as :
\begin{equation}
b = B(M, e) = \alpha(\beta - C(M, e))
\end{equation}
where $\alpha$ and $\beta$ are hyperparameters. After the bonus computation, the observation embedding is added to memory if the bonus b is larger than a novelty threshold $b_{novelty}$.

\section{Standard Testing Environments}

Through various methods reviewed above, almost all of them shared common environments in which these memory based modules had been tested. The first standard testing environment is the Arcade Learning Environment (Atari) \cite{mnih2013playing}. The Arcade Learning Environment is a suite of arcade games originally developed for the Atari-2600 console. These games are relatively simple visually but require complex and precise policies to achieve high expected reward, and form an interesting set of tasks as they contain diverse challenges such as sparse rewards and vastly different magnitudes of scores across games. This is a good choice as it acts like a baseline, as the most common algorithms like DQN \cite{mnih2013playing} and A3C \cite{mnih2016asynchronous} have been applied in this domain. 

The second common environment is a 2D and 3D maze-based environment, where memory is a crucial part for optimal planning. These mazes generally involve an agent to successfully navigate a 2D grid world populated with obstacles at random positions. The task is made harder by having random start and goal positions. These normally involve a partially observable markov decision process as the agent has only a single source of observation - the scene in front of it, while has no idea about the full map of the maze.

\begin{figure}[h]
    \centering
    \includegraphics[width=8cm]{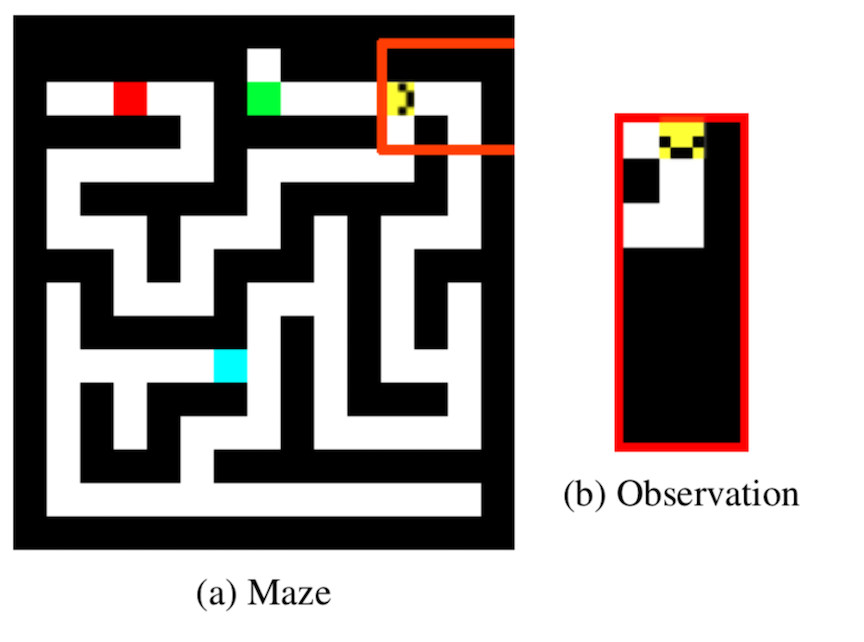}
    \caption{\textbf{2D Maze Environment} - The left side (a) represents the fully observable maze while the right side (b) represents the agent observations.}
    \label{fig:map}
\end{figure}

Generally, the 2D-mazes are generated using random generator. Hence, test set therefore represents maze geometries that have never been seen during training, and measure the agent’s ability to generalize to new environments. For testing in more complicated 3D environments, the 2D maze environment is implemented in 3D using the ViZDoom \cite{wydmuch2018vizdoom} environment and a random maze generator. In this environment, the indicator is a torch of either red or green color that is always at a fixed location in view of the player’s starting state. The goals are red/green towers that are randomly positioned throughout the maze. 
Other than these, some of the more recent approaches involve testing on the memory game of Concentration. The game is played with a deck of cards in which each card face appears twice. At the start of each game, the cards are arranged face down on a flat surface. A player’s turn consists of turning over any two of the cards. If their faces are found to match, the player wins those two cards and removes them from the table, then plays again. If the two cards do not match, the player turns them face down again, then play passes to the next player. The game proceeds until all cards have been matched and removed from the table. The winning strategy is to remember the locations of the cards as their faces are revealed, then use those memories to find matching pairs. The Concentration game is converted to a reinforcement learning environment using the Omniglot \cite{lake2015human} dataset. 

All of the above mentioned environments provide a benchmark for new upcoming reinforcement learning algorithms which involve external memory. Each of these environments tackle different aspect such as speed of learning, partial observability, long scale decision making etc. which shows the importance of memory in the future advancement of reinforcement learning.

\section{Conclusions}
In this paper we have presented a brief survey on memory based reinforcement learning. We focused on different memory modules and methods which enable episodic memory to be used for learning how to control and plan for an agent. We cover different methods which use these modules for different reinforcement learning based problems and give their advantages and disadvantages. We provide a brief but detailed insights to each of these methods and  cover the common testing environments which are normally used. This paper had been written to promote the idea of usage of external memory in reinforcement learning and provide insights on how these methods have been based on/adapted from the learning procedures which occur in the brain. This paper hopes to be a useful resource to provide a detailed overview of the field and to help in the future development of it.  

%\subsubsection*{Acknowledgments}

%Use unnumbered third level headings for the acknowledgments. All acknowledgments, including those to funding agencies, go at the end of the paper.

\bibliography{iclr2019_conference}
\bibliographystyle{iclr2019_conference}

\end{document}